\documentclass[letterpaper]{article} 

\usepackage[preprint]{aaai2027}
\usepackage[hyphens]{url}
\usepackage{graphicx}
\usepackage{natbib}
\usepackage{caption}
\usepackage{amsmath,amssymb}
\usepackage{booktabs}
\usepackage{tabularx}
\usepackage{array}
\usepackage{enumitem}
\usepackage{newfloat}

\DeclareCaptionStyle{ruled}{labelfont=normalfont,labelsep=colon,strut=off}
\DeclareFloatingEnvironment[fileext=loa,name=Algorithm,placement=tb]{algorithm}
\newcommand{\algindent}{\hspace*{1.15em}}
\newcommand{\algrule}{\vspace{1.5pt}\hrule\vspace{2.5pt}}
\newcommand{\algphase}[1]{\item[]\hspace*{-1.55em}%
  \textit{$\triangleright$~#1}}
\newcommand{\algcom}[1]{\hfill\textit{$\triangleright$~#1}}
\newenvironment{algsteps}
  {\begin{enumerate}[label=\footnotesize\arabic*:, align=right,
      leftmargin=1.85em, labelsep=0.4em, itemsep=0.75pt,
      topsep=2.5pt, parsep=0pt, partopsep=0pt]}
  {\end{enumerate}}

\newcommand{\kg}{\mathcal{G}}
\newcommand{\ctxstore}{\mathcal{C}}
\newcommand{\mem}{\mathcal{M}}

\title{Search Broadly, Seek Evidence on Both Sides, Decide Narrowly: Evidence-Admissible GraphRAG for Longitudinal Clinical Event Verification}
\author{
Xingtao Lin\textsuperscript{1},
Yubo Feng\textsuperscript{1},
Weixin Liu\textsuperscript{1},
Hangqi Ren\textsuperscript{1},
Junchao Zhou\textsuperscript{1},
Caiwan Sun\textsuperscript{1},
You Chen\textsuperscript{1,2}\corresponding
}
\affiliations{
\textsuperscript{1}Vanderbilt University\\
\textsuperscript{2}Vanderbilt University Medical Center (VUMC)
}

\begin{document}
\maketitle

\begin{abstract}
Longitudinal clinical event-relation verification determines whether a patient record supports a specified relation among two or more clinical events, such as whether acute kidney injury was documented after vancomycin within one encounter. Verification is difficult because the relevant information may be distributed across structured records, notes, laboratory trajectories, encounters, and time, while repeated documentation, negation, temporal mismatch, and conflicting findings can make retrieved information appear relevant without establishing the relation. We present MedEvent\-Graph-RAG, which represents individual event occurrences in a patient-specific graph and links each to its record source: structured rows, note spans, timestamps, local context, and numerical trajectories. A verification query specifies the events, the relation, and the clinical scope to be evaluated. The graph then guides retrieval of candidate event sequences connecting those events, together with source evidence that may support or contradict the relation, and redirects retrieval while required events or relations remain unresolved. Before assessment, the system excludes information belonging to another patient, outside the specified encounter-, episode-, or patient-level scope, or lacking a traceable source. A separate assessor evaluates only the resulting query-specific evidence bundle and returns supported, conflicting, refuted, or insufficient; graph scores, retrieval rankings, search history, and procedural memory guide retrieval but do not enter that bundle. Across ten protocols on i2b2, n2c2, MIMIC-IV, and LUNGUAGE, it reaches balanced accuracies of 78.6, 67.3, and 96.8 on pairwise temporal, medication--adverse-event, and recorded-order verification, exceeding the strongest matched baseline by 26.9, 4.9, and 30.4. When a required evidence component is removed it reaches 92.2 balanced accuracy with no false-support predictions; when intermediate events are withheld it returns complete event sequences with traceable evidence for every relation in 57.9\% of i2b2 and 70.0\% of LUNGUAGE cases. These results indicate that pairing patient event graphs with source-linked evidence can improve longitudinal verification and reduce unfounded conclusions when evidence is missing.
\end{abstract}

\section{Introduction}

Longitudinal clinical event-relation verification asks whether a patient's electronic health record supports a given statement about the relation between two or more clinical events. This task is challenging because electronic health records represent medications, laboratory measurements, diagnoses, procedures, and clinical assessments as separate observations across structured tables, narrative notes, encounters, and time, leaving clinically meaningful relations largely implicit~\citep{styler2014,bethard2016}. The required evidence may therefore form a chain linking medication administration, a subsequent creatinine rise, and a diagnostic assessment, while apparently relevant observations may be negated, copied forward, temporally misaligned, or drawn from another episode. Reliable verification consequently requires three capabilities: recovering the event occurrences needed to complete the chain, determining whether the recovered evidence supports or contradicts the specified relation, and grounding every event and relation in the correct patient, clinical scope, timestamp, and source.

These three capabilities fail for distinct reasons. Copied notes, templated
sections, and recurrent event types produce redundant mentions that exhaust a
limited retrieval budget while the occurrences needed to ground a
cross-encounter relation go unretrieved~\citep{wrenn2010}. Relevance does not
establish direction, because retrieved content may affirm, negate, qualify,
or retrospectively mention an event, and correctly identified events may
still be temporally incompatible with the queried relation or contradicted
elsewhere~\citep{negex,context2009,i2b2assertion}. Admissibility fails last:
a plausible chain may combine incompatible episodes, substitute documentation
time for occurrence time, or rest on copied statements whose originating
evidence cannot be resolved. Verification therefore needs broad signals to
discover candidate chains, but stricter criteria to decide from them.

We propose \textbf{MedEvent\-Graph-RAG}, an evidence-admissible framework
that searches broadly across longitudinal records, seeks evidence on both
sides of a verification query, and decides only from validated patient evidence. An
event-centric patient graph and a stateful controller discover
query-covering candidate chains across structured records, notes, numerical
trajectories, encounters, and time, linking every recovered occurrence to its
originating row, span, or trajectory and redirecting search while required
events, transitions, or contexts remain unresolved. The controller retrieves
relation-aligned and potentially contradictory observations alike, and a
separate assessor then evaluates support and refutation on independent axes,
distinguishing evidence on both sides from evidence on neither. A
query-specific evidence contract finally admits only patient-consistent,
scope-valid, and source-resolvable evidence, so graph scores, retrieval
priors, controller state, and procedural memory can influence what is
recovered but never what is concluded.

\begin{figure*}[t]
\centering
\includegraphics[width=\textwidth]{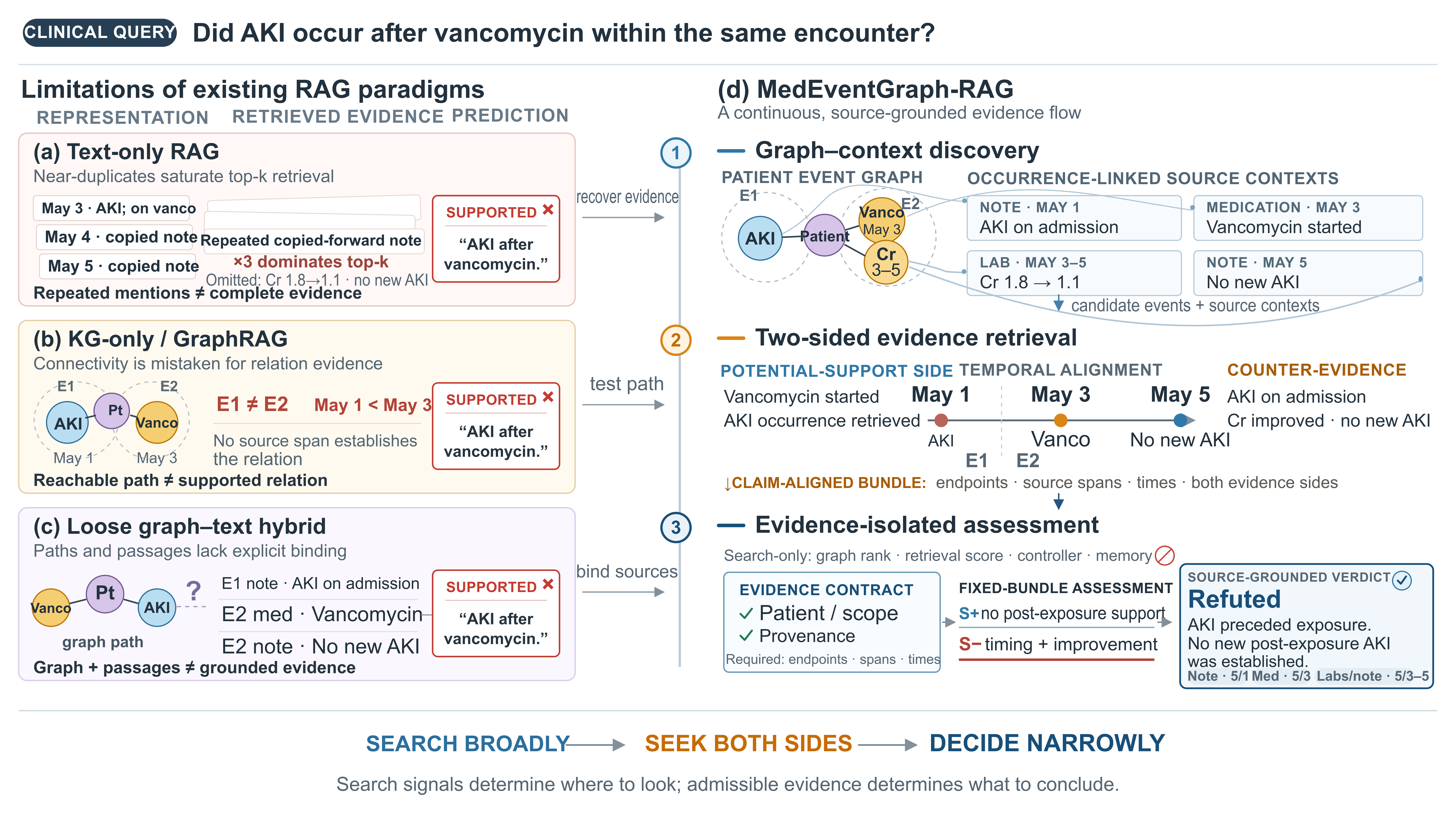}
\caption{MedEventGraph-RAG in one view. Existing RAG paradigms may retrieve
relevant but non-evidential content, whereas our framework binds event paths
to source contexts and isolates assessment from search signals.}
\label{fig:system-story}
\end{figure*}

\section{Related Work}

Graph-based and adaptive RAG broaden candidate discovery through relational expansion, graph--vector fusion, textual-subgraph retrieval, and iterative graph--document search~\citep{kg2rag,hipporag,hipporag2,lightrag,graphrag2024,raptor,grag,tog2,graphreader,gretriever,ggr,m3kgrag,selfrag,crag}. Clinical variants incorporate medical knowledge, patient history, temporal structure, and factual or counterfactual context for generation, prediction, and longitudinal retrieval~\citep{medgraphrag,medrag2025,kare,ehrrag,ehrragp}. These methods improve semantic access and structural reach, but do not separate the signals used to retrieve candidates from the evidence allowed to determine a patient-specific verdict. Domain-adapted clinical LLMs improve medical language understanding without enforcing this boundary~\citep{meditron,biomistral}.

Agent-memory methods reuse reflections or workflows to improve future behavior~\citep{reflexion,expel,awm,memgpt}, but instance-level memory could create an unintended cross-record channel in patient-specific verification. We instead retain only patient-independent retrieval policies.

\section{Method}

\begin{algorithm}[!tb]
\hrule\vspace{2pt}
\caption{Broad, two-sided discovery with evidence-admissible assessment.}
\label{alg:controller}
\vspace{2pt}\hrule\vspace{3pt}
\small
\textbf{Input:} query $q$; patient graph $\kg_p$; context store $\ctxstore_p$;
optional memory $\mem$; budgets $K,B,d_{\max},R_{\max}$\\[1pt]
\textbf{Output:} verdict $y$; evidence axes $(S^+,S^-)$; frozen bundle
$\mathcal{B}_c$
\algrule
\begin{algsteps}
\item $c\leftarrow\textsc{Parse}(q)$;\ \ $u\leftarrow\textsc{InitObligations}(c)$
\item $\pi\leftarrow\textsc{InitPolicy}(c,\mem)$;\ \ $r\leftarrow0$
\item $F_0\leftarrow\textsc{Anchors}(c,\kg_p)$;\ \ $\mathcal{I}\leftarrow\emptyset$
\item \textbf{for} $d=1$ \textbf{to} $d_{\max}$ \textbf{do}
\algphase{\algindent search broadly, on both sides}
\item \algindent $P_d\leftarrow\textsc{Expand}(F_{d-1},\kg_p;K,\pi)$
\item \algindent $(C_d^{+},C_d^{-})\leftarrow
      \textsc{TwoSidedRetrieve}(P_d,c,u;\ctxstore_p)$
\item \algindent $(P_d,C_d)\leftarrow\textsc{BindDedup}(P_d,C_d^{+}\cup C_d^{-})$
\item \algindent $(P_d,u)\leftarrow\textsc{CandidateCheck}(P_d,C_d,c,u)$
\item \algindent $\widehat{P}_d\leftarrow\textsc{TopB}(P_d;R(\cdot\mid c),B)$
\algphase{\algindent decide narrowly, from admissible evidence only}
\item \algindent \textbf{for each} $P\in\widehat{P}_d$ \textbf{do}
\item \algindent\algindent $\mathcal{I}\leftarrow\textsc{SourceDedup}
      \bigl(\mathcal{I}\cup\textsc{Contract}(c,P,C_d(P))\bigr)$
\item \algindent $\mathcal{B}_c\leftarrow(c,\mathcal{I})$
      \algcom{excludes $\pi,R,\mem$}
\item \algindent $(S^{+},S^{-},y)\leftarrow\textsc{Assess}_{\theta}(\mathcal{B}_c)$
\item \algindent $u\leftarrow\textsc{UpdateObligations}(u,c,\mathcal{B}_c)$
\item \algindent \textbf{if} $y=\textit{conflicting}$ \textbf{or}
      $\textsc{Resolved}(u,c)$ \textbf{then break}
\algphase{\algindent replan retrieval only, never assessment}
\item \algindent $f\leftarrow\textsc{StructuredFailure}(u,c)$
\item \algindent \textbf{if} $f\neq\emptyset$ \textbf{and} $r<R_{\max}$
      \textbf{then}
\item \algindent\algindent $\pi\leftarrow\textsc{Replan}(\pi,f,\mem)$;\ \ 
      $r\leftarrow r+1$
\item \algindent $F_d\leftarrow\textsc{Frontier}(P_d,u,\pi)$
\item \textbf{return} $y$, $(S^{+},S^{-})$, $\mathcal{B}_c$
\end{algsteps}
\vspace{2pt}\hrule
\end{algorithm}

\begin{figure*}[t]
\centering
\includegraphics[width=\textwidth]{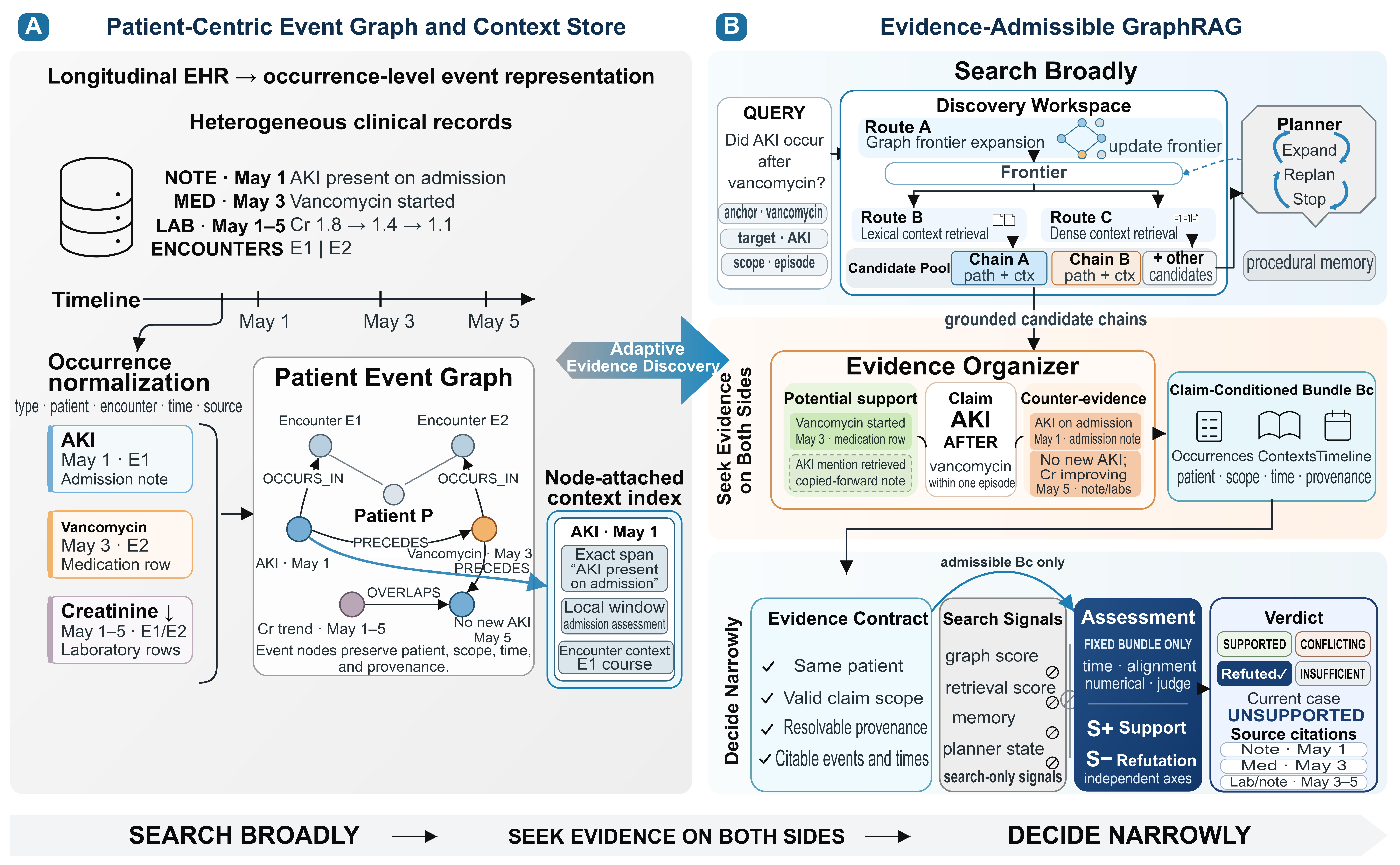}
\caption{Patient event graphs index source-resolvable contexts; adaptive
two-sided discovery builds a query-specific bundle, and only admissible
evidence reaches the fixed-bundle assessor.}
\label{fig:overview}
\end{figure*}

\paragraph{Task and information-flow invariants.}
Given a record $p$ and query $q$, we parse a normalized verification query $c=(E,R,\phi)$, where $E$ holds observed and unresolved event slots, $R$ their relation constraints, and $\phi\in\{\textit{encounter},\textit{episode},\textit{patient}\}$ the clinical scope. Unspecified fields stay unresolved. The system returns
$y\in\{\textit{supported},\textit{conflicting},
\textit{refuted},\textit{insufficient}\}$
with its source-resolvable evidence and citations. The information flow is
\begin{align}
D_c &= \textsc{Discover}(c,\kg_p,\ctxstore_p;\pi), \nonumber\\
\mathcal{B}_c &= \textsc{Contract}(c,D_c), \nonumber\\
(S^+,S^-,y) &= \textsc{Assess}(c,\mathcal{B}_c),
\label{eq:information-flow}
\end{align}
where $\kg_p$ and $\ctxstore_p$ are the patient event graph and source-linked
context store, $\pi$ is the retrieval policy, $D_c$ includes candidates
and search metadata, and $\mathcal{B}_c$ contains only admitted evidence.
Connectivity and retrieval relevance nominate candidates but do not establish the relation, and the direction in which evidence was retrieved does not fix its evidential polarity. Every item in $\mathcal{B}_c$ must be patient-consistent, occurrence-bound, source-resolvable, and not scope-invalid, while graph scores, retrieval ranks, controller state, search history, and memory remain outside $\mathcal{B}_c$ and cannot alter the assessment of an identical bundle.

\paragraph{Search broadly: patient event--context representation.}
Each record is $(\kg_p,\ctxstore_p)$, separating structural navigation from
evidential interpretation. $\kg_p$ indexes occurrences with their recorded
encounter and temporal structure; its edges are limited to recorded relations
and deterministic derivations, never gold relations, assessor judgments, or
model-inferred relations. $\ctxstore_p$ maps each occurrence to its originating structured row, narrative span, or numerical trajectory, preserving negation, uncertainty, historical status, values, timestamps, and sources. Thus $\kg_p$ broadens discovery, while the source-linked contexts supply the evidence from which relations are actually assessed.

\paragraph{Search broadly and seek evidence on both sides: adaptive discovery.}
Given the normalized query $c$, discovery initializes a patient-scoped frontier and iteratively expands candidate event chains across time and encounters. At depth $d$ the controller traverses structurally valid neighbors without requiring agreement with the specified relation, retrieving their occurrence-linked contexts through lexical, dense, and metadata-aware channels. Retrieval maintains complementary obligations: aligned queries seek observations consistent with the specified events and relations, while counter-oriented queries target negation, incompatible order, historical or hypothetical mentions, stable findings, and conflicting documentation. Retrieved contexts are bound to candidate occurrences, deduplicated, and annotated with patient, scope, timestamp, and source status. Malformed and cross-patient candidates are removed, while temporal or type patterns incompatible with the specified relation stay eligible, because they are exactly where counter-evidence tends to appear.

Because only a bounded number of candidates can be contracted, each path $P$ receives a query-conditioned priority score
\begin{equation}
R(P\mid c)
= \prod_{k\in\mathcal{K}(c)}
\left[\epsilon+(1-\epsilon)s_k(P,c)\right],
\label{eq:path-rank}
\end{equation}
where $\epsilon=0.2$ and each $s_k(P,c)\in[0,1]$ is a normalized discovery signal measuring graph diffusion, path compactness, source-context relevance, source-link completeness, or temporal and event-type compatibility. For a compositional query $c=(e_0,r_1,e_1,\ldots,r_h,e_h)$, $\mathcal{K}(c)$ additionally includes transition-evidence coverage
\begin{equation}
s_{\mathrm{cov}}(P,c)
= \frac{1}{h}
\sum_{i=1}^{h}
\mathbb{I}\!\left[
\Gamma_P(e_{i-1},r_i,e_i)\neq\emptyset
\right],
\label{eq:transition-coverage}
\end{equation}
where $\Gamma_P(e_{i-1},r_i,e_i)$ denotes the source-linked contexts, timestamps, or trajectories available to assess transition $(e_{i-1},r_i,e_i)$ on $P$. This measures whether each transition can be examined, not whether it supports the specified relation, so disagreement does not by itself eliminate a path. The lifted product maps every component to $[\epsilon,1]$, so weak signals lower priority without acting as hard filters.

At each depth the highest-priority candidates are contracted and assessed after source-level deduplication; path rank and depth are not aggregation weights. If an obligation remains unresolved the controller revises its policy from the corresponding structured failure and continues, stopping when both obligations resolve, admissible evidence establishes conflict, or the budget is exhausted. In Algorithm~\ref{alg:controller}, $d_{\max}$ bounds event-to-event traversal and $R_{\max}$ bounds policy revisions.

\paragraph{Decide narrowly: evidence contract.}
Discovery output is converted into a frozen, query-specific evidence bundle
\begin{equation}
\mathcal{B}_c=(c,\mathcal{I}_c),
\qquad
\mathcal{I}_c=
\{z_j=(E_j,C_j,T_j,\Pi_j)\}_{j=1}^{m},
\label{eq:evidence-contract}
\end{equation}
where each item $z_j$ binds one or more occurrences $E_j$ to their source contexts $C_j$, recorded time types and granularities $T_j$, and structured-row or note-span provenance $\Pi_j$, keeping contexts, timestamps, and citations attached to what they ground. An item $z$ is admitted only if
\begin{align}
\operatorname{Adm}(z;c) &=
A_{\mathrm{patient}}(z)
\land A_{\mathrm{binding}}(z)
\land A_{\mathrm{source}}(z) \nonumber\\
&\quad \land \bigl[A_{\mathrm{scope}}(z;c)\neq\mathrm{invalid}\bigr],
\label{eq:admissibility}
\end{align}
where $A_{\mathrm{scope}}(z;c)\in
\{\mathrm{valid},\mathrm{unknown},\mathrm{invalid}\}$.
Every admitted item must therefore resolve to the target patient, to specific occurrences, and to original sources, and items used jointly must be consistent with the query-defined scope. Missing episode metadata or coarse timing stays explicitly \emph{unknown}: retained for assessment, but unable by itself to establish support or refutation. The contract enforces patient identity, scope consistency, occurrence--source binding, and source resolution at assessment time; it does not guarantee correct event extraction, clinical interpretation, or the factual accuracy of the underlying record.

\paragraph{Decide narrowly: isolated two-sided assessment.}
After source-level deduplication, the assessor receives only the frozen bundle $\mathcal{B}_c$. It first assigns $v_{\mathrm{tim}}\in\{\textit{compatible},\textit{incompatible},\textit{unknown}\}$ from recorded times, granularities, and source-linked spans; coarse times remain \textit{unknown}, and an incompatible order contributes to refutation only when occurrences, scope, timestamps, and sources are jointly resolved. Graph-native path order carries no evidential force of its own. It then scores admissible evidence for and against the specified relation on independent axes:
\begin{align}
S^{+} &= \alpha_{\mathrm{tim}}S_{\mathrm{tim}}^{+}
+ \alpha_{\mathrm{num}}S_{\mathrm{num}}^{+}
+ \alpha_{\mathrm{jdg}}S_{\mathrm{jdg}}^{+},
\nonumber\\
S^{-} &= S_{\mathrm{jdg}}^{-},
\label{eq:evidence-axes}
\end{align}
where $S_{\mathrm{tim}}^{+}$ measures source-grounded temporal compatibility, $S_{\mathrm{num}}^{+}$ numerical evidence, and $S_{\mathrm{jdg}}^{\pm}$ assessor-identified supporting and refuting evidence. The refutation axis is conservative: incompatibility raises $S^{-}$ only on explicit, source-grounded counter-evidence, each citing the item grounding it. The axes are evidence strengths, not complementary probabilities, so evidence on both sides raises both while unresolved evidence raises neither. The weights are development-split operating points, frozen at $\alpha_{\mathrm{tim}}{=}0.45$, $\alpha_{\mathrm{num}}{=}0.22$, and $\alpha_{\mathrm{jdg}}{=}0.33$, and are never re-calibrated per dataset or per protocol. Thresholds selected on the same split map the two axes to the verdict:
\begin{equation}
y=
\begin{cases}
\mathrm{supported},
& S^{+}\geq\tau_{+},\; S^{-}<\tau_{-},\\
\mathrm{conflicting},
& S^{+}\geq\tau_{+},\; S^{-}\geq\tau_{-},\\
\mathrm{refuted},
& S^{+}<\tau_{+},\; S^{-}\geq\tau_{-},\\
\mathrm{insufficient},
& S^{+}<\tau_{+},\; S^{-}<\tau_{-}.
\end{cases}
\label{eq:quadrants}
\end{equation}
For binary benchmarks the three non-supported verdicts are collapsed after four-way inference and solely for scoring, retaining verdict, axes, citations, and failure type for audit. With model revision, prompt, and decoding fixed, assessment satisfies fixed-bundle independence:
\begin{equation}
\mathcal{B}_c^{(1)}=\mathcal{B}_c^{(2)}
\Longrightarrow
\textsc{Assess}_{\theta}(\mathcal{B}_c^{(1)})
= \textsc{Assess}_{\theta}(\mathcal{B}_c^{(2)}).
\label{eq:fixed-bundle}
\end{equation}

\paragraph{Compositional and hidden-intermediate queries.}
We extend pairwise verification to compositional queries $\chi=(e_0,r_1,e_1,\ldots,r_h,e_h)$ with $h\geq2$, where each $e_i$ is a required occurrence and each $r_i$ the relation between $e_{i-1}$ and $e_i$. Such a query is supported only if a scope-consistent chain grounds every occurrence and supplies admissible evidence for every adjacent relation; recovering the nodes or their graph order is insufficient, since each transition must be assessed from its bound textual, numerical, or temporal evidence.

The hidden-intermediate setting replaces $\chi$ with $q_{\mathrm{hid}}=(e_0,e_h,\rho_{1:h},\tau_{1:h-1})$, withholding the intermediate identities and leaving only endpoints, relation sequence, and type constraints. We therefore separate structural recovery, which identifies a candidate chain, from evidential verification, which additionally requires admissible evidence and citations for every relation.

\paragraph{Retrieval-only procedural memory.}
As an optional extension, $\mem$ is a frozen bank of patient-independent
retrieval policies derived from recurring training and development search
failures; it holds no patient identifiers or facts, event values, evidence
spans, provenance, labels, answers, or verdicts. Given a query and structured failure $f$, the controller may retrieve
$\pi=\textsc{RetrievePolicy}(\mem,c,f)$ to revise expansion, query selection,
or replanning. Memory therefore changes only which admissible
evidence is recovered, leaving Eq.~\eqref{eq:fixed-bundle} intact.

\section{Task Construction}
\label{sec:tasks}

Existing clinical resources annotate relations or answer questions; none
scores whether a record supplies admissible evidence for a specified relation. We
therefore construct ten protocols from three annotated corpora and state
their sources, negatives, and leakage controls here.

\paragraph{Query sources and labels.}
Pairwise queries come from i2b2 2012 TLINK annotations (T1), n2c2 2018
medication--ADE annotations (T2), and recorded event order in MIMIC-IV
structured tables (T3)~\citep{i2b2,n2c2ade,mimiciv,mimicnote}. In each case
the annotation supplies only the query endpoints and the held-out label.
Relation direction never enters events, edges, contexts, prompts, retrieval
scores, or memory, so no method can read a verdict from the artifact it
searches. A release audit verifies this before evaluation: the selectable edge set
contains no support relation, gold direction appears only in evaluation
labels, positive and negative candidates share one graph schema, and
train/dev/test patients do not overlap.

\paragraph{Negatives.}
A negative differing structurally from a positive can be solved by surface
statistics rather than evidence, so each negative is a relation perturbation
of a positive chain over the same events and contexts.
Measured from the protocol alone, with no model involved, positives and
negatives on T9 i2b2 have chain lengths 4.05 and 3.90, hops 3.05 and 2.90, and
8.21 and 8.74 contexts per queried endpoint; on T9 n2c2 the classes are
identical in both. Neither carries a retrieval-side signature.

\paragraph{The ten protocols.}
T1--T3 are the pairwise tasks defined above: given two named events, decide
whether the record supports, refutes, conflicts on, or is insufficient for
the stated relation. The other seven isolate one capability each.
\textbf{T4} scores retrieval alone, asking which node-attached contexts a
method recovers before any verdict is formed. \textbf{T5} masks the event, context, provenance, or timing that would
establish a supported verdict, so the correct answer becomes insufficient or
conflicting and a method still answering supported is answering from
relevance. \textbf{T6} leaves the evidence intact but adds four distractor strata:
other-patient events, other-patient events sharing a name, other-patient
contexts carrying the exact query terms, and same-patient contexts from a
query-incompatible encounter. \textbf{T7} runs the system with and without procedural
memory on identical identifiers. \textbf{T8} is a patient-disjoint
latest-time split testing transfer rather than memorization. \textbf{T9}
states every queried event and asks whether all adjacent relations hold
jointly, so one refuted transition refutes the query. \textbf{T10} withholds the intermediate identities and scores whether a
reconstructed chain is also verified and source-bound; it additionally runs on
LUNGUAGE~\citep{lunguage}.

\paragraph{What is held fixed.}
Every method receives the same frozen artifacts, identifiers, patient filters,
encoders, candidate caps, context budget, and assessor; only the online
retrieval operator differs. No protocol is filtered by any system's output.

\section{Experiments}
\label{sec:experiments}

\subsection{Setup}

\paragraph{Baselines and implementation.}
B1 and B2 are direct and text-only retrieval; B3--B5 add the patient graph, a
loose hybrid, and coupled graph--context retrieval without isolated
assessment; B6 is the full system and B7 adds procedural memory. We reproduce the online retrieval operators of MedGraphRAG, EHR-RAG,
HippoRAG~2, KG$^2$RAG, LightRAG, and GRAG on the shared index rather than
importing published values. Qwen3-8B performs parsing and control, and the
development-selected Qwen3-32B assesses frozen bundles~\citep{qwen3};
defaults are $d_{\max}{=}R_{\max}{=}5$ and 12 final contexts.

\begin{table*}[!tb]
\centering
\small
\textbf{(a) Primary results}\\[2pt]
\begin{tabular}{@{}l@{\hspace{4pt}}c@{\hspace{4pt}}c@{\hspace{4pt}}c@{\hspace{4pt}}c@{\hspace{4pt}}c@{\hspace{6pt}}c@{\hspace{4pt}}c@{\hspace{4pt}}c@{\hspace{4pt}}c@{}}
\toprule
& \multicolumn{5}{c@{\hspace{9pt}}}{Discrimination and evidence recovery}
& \multicolumn{4}{c}{Evidence integrity} \\
\cmidrule(r{9pt}){2-6}\cmidrule{7-10}
Method & T1 & T2 & T3 & Ctx P@10$^{1/2/3}_{\;\ddagger}$ & Pair R$^{1/2/3}$
& T5 & T6 & FSR$\downarrow$ & Sup.-rec \\
\midrule
B1 Direct LLM & 46.1/33.0 & 62.4/62.4 & 50.2/40.4 & -- & -- & 46.6/42.6 & 48.1/33.6 & \textbf{11.5} & 7.6 \\
B2 Text-RAG & 50.0/37.5 & 51.3/36.2 & 53.0/47.3 & -- & -- & 21.0/29.4 & 48.3/39.2 & 96.6 & 93.1 \\
B3 KG-only & 47.8/44.3 & 48.7/48.5 & 66.4/63.2 & -- & -- & 43.0/41.6 & 50.3/42.5 & 20.7 & 21.4 \\
B4 Loose hybrid & 51.7/49.9 & 50.4/49.7 & 62.6/62.3 & -- & -- & 43.9/54.9 & 47.8/47.4 & 70.1 & 65.6 \\
B5 Graph--text & 50.1/35.9 & 39.4/33.9 & 50.2/35.1 & -- & -- & 20.8/29.0 & 51.9/43.6 & 93.1 & \textbf{96.9} \\
B6 Isolated & \textbf{78.6/78.5} & \textbf{67.3/65.7} & \textbf{96.8/96.8} & \textbf{29.5/27.3/19.4} & \textbf{68.5/53.5/71.6} & \textbf{92.2/94.4} & 75.5/75.0 & 25.3 & 76.3 \\
B7 Full & \textbf{78.6/78.5} & \textbf{67.3/65.7} & \textbf{96.8/96.8}$^\dagger$ & \textbf{29.5/27.3/19.4} & \textbf{68.5/53.5/71.6} & \textbf{92.2/94.4} & \textbf{76.7/76.3} & 25.3 & 78.6 \\
\bottomrule
\end{tabular}

\medskip
\textbf{(b) Patient-cluster inference}\\[2pt]
\begin{tabular}{@{}lccclcc@{}}
\toprule
Protocol & B6 BA [95\% CI] & B6 F1 [95\% CI] & Strongest B1--B5 & $\Delta$BA
& Holm $p_{\mathrm{BA}}$ & Holm $p_{\mathrm{F1}}$ \\
\midrule
T1 i2b2 & 78.6 [75.4, 81.8] & 78.5 [75.2, 81.7] & B4 (51.7) & +26.9 & .0074 & .0074 \\
T2 n2c2$^\ddagger$ & 67.3 [62.0, 72.8] & 65.7 [59.2, 72.2] & B1 (62.4) & +4.9 & 1.000 & 1.000 \\
T3 MIMIC & 96.8 [95.2, 98.2] & 96.8 [95.2, 98.2] & B3 (66.4) & +30.4 & .0074 & .0074 \\
T5 masked & 92.2 [87.7, 96.5] & 94.4 [91.0, 97.5] & B1 (46.6) & +45.6 & .0074 & .0074 \\
T6 scope & 75.5 [64.1, 87.4] & 75.0 [63.3, 86.2] & B5 (51.9) & +23.6 & .0130 & .0074 \\
T9 MIMIC & 61.1 [57.6, 65.1] & 59.8 [56.0, 64.1] & B5 (51.9) & +9.2 & .0074 & .0074 \\
\bottomrule
\end{tabular}
\caption{Percent BA/F1 unless noted; superscripts 1/2/3 index T1/T2/T3, and
FSR and supported recall are from T6. Bold marks the best value in a column,
including where a control attains it: B1 has the lowest FSR and B5 the highest
supported recall, each at the other's expense. Panel (b) gives 95\% bootstrap
intervals and Holm-adjusted $p$-values. $^\dagger$B7 equals B6 on T3 by
construction; $^\ddagger$exploratory. $_{\ddagger}$Ctx P@10 has a fixed
denominator of 10 while a query has only 5.11, 3.18, and 2.85 gold contexts on
average, capping it at 49.1, 31.8, and 28.5; the reported values reach 60\%,
86\%, and 68\% of that cap.}
\label{tab:main}
\end{table*}

\begin{table}[!tb]
\centering
\small
\begin{tabular}{@{}lc@{}}
\toprule
Method & T1/T2/T3 BA \\
\midrule
MedGraphRAG & 72.3/54.4/78.8 \\
EHR-RAG & 56.5/56.6/70.0 \\
HippoRAG~2 & 57.9/61.9/64.8 \\
KG$^2$RAG & 61.0/56.6/70.2 \\
LightRAG & 71.3/52.2/76.8 \\
GRAG & 57.7/56.2/68.4 \\
\textbf{MedEventGraph-RAG} & \textbf{78.6/67.3/96.8} \\
\bottomrule
\end{tabular}
\caption{Matched-index retrieval. Artifacts, identifiers, context budget, and
assessor are held fixed and only the online retrieval operator changes; bold
marks the best value per task. Inference cost is reported in the supplement.}
\label{tab:external}
\end{table}

\paragraph{Metrics.}
Balanced accuracy (BA) and macro-F1 are primary for labeled protocols. The
false support rate (FSR) is always reported together with supported recall,
so that indiscriminate abstention cannot appear safe. T10 reports context recall (Ctx-R), hidden-bridge recall (Bridge),
complete-chain recall at 20 (C@20), verified-chain recall (Verif), and
provenance-valid chain recall (Prov.). Confidence intervals use 10,000 patient-cluster bootstrap replicates, and
paired randomization tests use 10,000 permutations with Holm correction over
the complete comparison family. T9 i2b2, T9 n2c2, and T10 n2c2 ($n{=}11$) are
exploratory because of their small patient-cluster counts.

\begin{table}[!tb]
\centering
\small
\begin{tabular}{@{}l@{\hspace{3pt}}c@{\hspace{3pt}}c@{\hspace{3pt}}c@{\hspace{3pt}}c@{\hspace{3pt}}c@{}}
\toprule
Stage & BA & F1 & FSR$\downarrow$ & Ctx-R & Chain-R \\
\midrule
E1 bound graph--context & 70.0 & 69.9 & 23.3 & 46.0 & 52.5 \\
E2 + two-sided & 72.5 & 72.2 & 16.7 & 63.1 & 47.5 \\
E3 + isolated axes & 75.0 & 74.7 & 15.0 & 63.1 & 47.5 \\
E4 + replanning & 75.8 & 75.7 & 15.8 & 60.0 & 55.0 \\
E5 + memory & 76.2 & 76.1 & 15.8 & 60.0 & 55.0 \\
\bottomrule
\end{tabular}
\caption{Cumulative ablation, percent, on a frozen balanced 240-query set;
Chain-R is complete-chain recall. Stages add search over the patient graph with its
occurrence-bound context index (E1), source-bound two-sided retrieval (E2),
evidence-isolated dual-axis verification (E3), verifier-guided replanning
(E4, our B6), and retrieval-only procedural memory (E5, our B7), under the
same candidate contract, assessor, and retrieval budget. E1--E3 decide from a coupled
boundary and E4--E5 from an isolated one; E1--E3 hold the candidate superset
fixed, while E4--E5 may extend it because that adaptive change is the
treatment measured.}
\label{tab:staged-ablation}
\end{table}

\begin{table}[!tb]
\centering
\small
\textbf{(a) T9: all events named}\\[2pt]
\begin{tabular}{@{}l@{\hspace{2pt}}c@{\hspace{2pt}}c@{\hspace{2pt}}c@{}}
\toprule
Method & i2b2 & n2c2 & MIMIC \\
\midrule
B2 Text-RAG & 52.5/42.0 & 50.0/33.3 & 50.2/49.5 \\
B3 KG-only & 47.5/32.2 & 50.7/50.7 & 50.3/34.9 \\
B4 Loose hybrid & 67.5/67.0 & 47.9/35.6 & 49.4/42.8 \\
B5 Graph--text & 50.0/33.3 & 50.0/33.3 & 51.9/51.9 \\
\textbf{Ours} & \textbf{82.5/82.2} & \textbf{66.4/65.3} & \textbf{61.1/59.8} \\
\bottomrule
\end{tabular}

\medskip
\textbf{(b) T10: intermediates withheld}\\[2pt]
\begin{tabular}{@{}lccccc@{}}
\toprule
Method & Ctx-R & Bridge & C@20 & Verif & Prov. \\
\midrule
B1 & n/a & n/a & n/a & n/a & n/a \\
B2 & 59.6 & n/a & n/a & n/a & n/a \\
B3 & n/a & 56.1 & 5.8 & n/a & n/a \\
B4 & 59.6 & 56.1 & 5.8 & n/a & n/a \\
B5 & 50.2 & 90.1 & 67.1 & n/a & n/a \\
\textbf{B6} & \textbf{65.2} & \textbf{99.9} & \textbf{77.1} & \textbf{77.9} & 57.9 \\
\textbf{B7} & \textbf{65.2} & \textbf{99.9} & \textbf{77.1} & \textbf{77.9} & \textbf{58.8} \\
\midrule
Ours-L & 92.3 & 98.8 & 77.1 & 80.7 & 70.0 \\
\bottomrule
\end{tabular}
\caption{Compositional queries. (a) Percent BA/F1 on debiased balanced
splits ($n{=}40/140/620$). (b) Evidence survival from context and bridge
recovery through complete candidate (C@20), verification, and provenance.
Each metric in (b) is defined only for architectures that emit the
corresponding output, and n/a marks a stage a method lacks: B1 retrieves
nothing, B2 builds no graph, B3 attaches no contexts, and no B1--B5 variant
emits a verification path or a source-bound chain. We write n/a rather than 0
because the quantity is undefined for them, not measured at zero, and claim
nothing about an extended variant. Bold marks the best among B1--B7; Ours-L
is LUNGUAGE, listed below the rule rather than compared in-column.}
\label{tab:compositional}
\end{table}

\subsection{Results}

Four tables support the analysis, each controlling a different confound.
Table~\ref{tab:main} reports the pairwise protocols T1--T3 together with the
evidence-integrity protocols T5 and T6, placing accuracy beside a
false-support column so that neither can be raised alone.
Table~\ref{tab:external} asks whether any gain is an artifact of our index
rather than our search policy, so it holds artifacts, identifiers, context
budget, and assessor fixed and replaces only the online retrieval operator
with that of six published systems, reproduced on the shared index rather
than imported from their papers. Table~\ref{tab:staged-ablation} attributes
the gain internally by adding one evidence-layer mechanism at a time over a
frozen balanced 240-query set, so that a change in what is retrieved can be
separated from a change in how it is judged. Table~\ref{tab:compositional} moves from pairwise to compositional queries,
with every event named in panel (a) and the intermediate identities withheld
in panel (b).

\paragraph{Does broad, source-linked discovery recover evidence that flat and graph-only retrieval miss?}
B6 gains 26.9, 4.9, and 30.4 BA over the strongest matched control
(Table~\ref{tab:main}). Patient-cluster inference confirms T1 and T3 after
Holm correction ($p{=}.0074$); T2 is exploratory, and T3 measures recorded
order rather than causality. Under the matched index the best external result per task is 72.3, 61.9, and
78.8, attained by three different systems, against our 78.6, 67.3, and 96.8
(Table~\ref{tab:external}), so what differs is the search policy rather than
the index. Two measurements bound the result: T4 reaches 89.5 context recall, so the
occurrence-linked contexts are retrievable, and T8 retains 71.5/69.5 on a
patient-disjoint split, so the gain survives distribution shift.

\paragraph{Does relevance establish evidential direction?}
It does not, and balanced accuracy alone conceals this: under T5 a system
that answers from relevance must produce false support. The retrieval-coupled
controls B2 and B5 do exactly that, reporting false support on 96.6\% and
93.1\% of non-supported queries: they answer \emph{supported} almost whenever
retrieval returns anything. The opposite
failure is equally available, since B1 holds FSR to 11.5 only by abstaining
almost everywhere, at 7.6\% supported recall. B6 is the only configuration that separates
the two axes, reaching 92.2/94.4 BA/F1 with zero false support on T5
($+45.6$ BA over B1, $p{=}.0074$) while holding FSR at 25.3 with 76.3\%
supported recall overall. One case shows the mechanism: asked whether kidney transplantation preceded
pancreas transplantation once the establishing context was masked, B5
returned \emph{supported} from residual graph linkage, while B6 returned
\emph{insufficient} and named the missing provenance. The graph proposes the pair; it cannot substitute for the source
that orders it.

\paragraph{Does admissibility buy safety by discarding evidence?}
It does not. Under T6 B6 reaches 75.5/75.0
($+23.6$ BA over B5, $p{=}.0130$) while keeping 76.3\% supported recall, so
the patient boundary holds without rejecting legitimate cross-encounter
evidence; blanket rejection would have depressed both quantities. The cumulative ablation (Table~\ref{tab:staged-ablation}) separates the two
routes to that result. E2 acts on what is found: context recall rises from 46.0 to 63.1 and FSR
falls from 23.3 to 16.7. E3 then acts only on how that evidence is judged,
leaving context and chain recall unchanged at 63.1 and 47.5 while FSR falls to
15.0 and BA rises to 75.0.
Withholding an unfounded decision and failing to
retrieve evidence are thus separable, and here the boundary costs no
accuracy. Verifier-guided replanning (E4) then lifts chain recall to 55.0 at 75.8 BA,
so the boundary does not foreclose further search. Retrieval-only procedural
memory (E5) adds 0.4 BA/F1 at unchanged FSR (15.8), context recall (60.0), and
chain recall (55.0).

\paragraph{Where does the compositional pipeline still fail?}
T9 exceeds the strongest B2 to B5 control on every corpus
(Table~\ref{tab:compositional}a), though only the larger MIMIC split supports a corrected
result ($+9.2$ BA, $p{=}.0074$). T10 explains why
(Table~\ref{tab:compositional}b). Read as a funnel from bridge recovery through complete
candidate at 20, verification, and provenance, the comparison is one of
capability before degree. B5 reaches 90.1 and 67.1, so structural reach is attainable with no grounding
stage at all. The substantive result is that our system populates the last two
columns at all, and non-trivially: B6 lifts the funnel to 99.9 and 77.1,
preserves 77.9 through verification, yet retains only 57.9 as
provenance-valid. Structural reach is therefore solved while source binding is not, and the 20.0-point drop from verification to
provenance localizes the bottleneck. Ours-L shows the bottleneck is contextual rather than architectural: richer
contexts raise Ctx-R from 65.2 to 92.3 and provenance to 70.0 at unchanged
C@20.

\paragraph{Can procedural memory improve retrieval without reaching judgment?}
B7 stores no patient facts, spans, answers, or verdicts. It improves T6 by
1.2/1.3 BA/F1 through three corrected retrieval failures at unchanged FSR
(25.3), so the gain enters through what is found rather than how it is judged,
and equals B6 exactly on T1--T3 and T5. Fixed-bundle replay with memory disabled, enabled,
or injected with fact-like content changed no verdict and no $(S^{+},S^{-})$.

\paragraph{What does the boundary cost?}
Evidence isolation is materially more expensive: B6 uses 4.625 calls and
12.8k tokens per query against 1.000 and 1.4k for B2, a $4.6\times$ call and
$9.3\times$ token premium, while procedural memory adds 3.06 seconds of local
work and no LLM call. Per-method figures are in the supplement.

\FloatBarrier
\section{Discussion and Conclusion}

Longitudinal verification fails in three independent ways, each answered by a
design decision rather than by scale. Binding the unit of search to an event
occurrence and its source steers discovery by what still needs grounding,
lifting hidden-bridge recovery to 99.9; scoring support and refutation on
independent axes yields zero false support under masked evidence, against 96.6
and 93.1 for the coupled controls; and a contract applied at assessment time
holds the patient boundary under distractors while retaining 76.3\% supported
recall. These answers are separable, and the residual gap is
localized: we verify 77.9 of reconstructed chains but ground 57.9. The
principle generalizes beyond medicine: a verdict-producing system should
separate the signals deciding where to look from the evidence permitted to
decide, though we claim neither causality nor clinical safety beyond the
corpora tested.

\bibliography{references}

\end{document}